\documentclass[runningheads]{llncs}
\usepackage{graphicx}

\usepackage{tikz}
\usepackage{comment}
\usepackage{amsmath,amssymb} 
\usepackage{color}
\usepackage{multirow}
\usepackage[accsupp]{axessibility}  

\begin{document}
\pagestyle{headings}
\mainmatter
\def\ECCVSubNumber{259}  

\title{Meta-Causal Feature Learning for Out-of-Distribution Generalization} 

\titlerunning{Meta-Causal Feature Learning for Out-of-Distribution Generalization}
%
\author{Yuqing Wang \and
Xiangxian Li \and
Zhuang Qi \and
Jingyu Li \and
Xuelong Li \and \\
Xiangxu Meng \and
Lei Meng \thanks{Corresponding author
}}
\authorrunning{W. Yuqing et al.}
%
\institute{Shandong University, Jinan, Shandong, China
\email {\{wang\_yuqing,xiangxian\_lee,jingyu\_lee\}@mail.sdu.edu.cn},
\email {97qizhuang@gmail.com},
\email {\{lixuelong,mxx,lmeng\}@sdu.edu.cn}
}
\maketitle

\begin{abstract}
Causal inference has become a powerful tool to handle the out-of-distribution (OOD) generalization problem, which aims to extract the invariant features. However, conventional methods apply causal learners from multiple data splits, which may incur biased representation learning from imbalanced data distributions and difficulty in invariant feature learning from heterogeneous sources. To address these issues, this paper presents a balanced meta-causal learner (BMCL), which includes a balanced task generation module (BTG) and a meta-causal feature learning module (MCFL). Specifically, the BTG module learns to generate balanced subsets by a self-learned partitioning algorithm with constraints on the proportions of sample classes and contexts. The MCFL module trains a meta-learner adapted to different distributions. Experiments conducted on NICO++ dataset verified that BMCL effectively identifies the class-invariant visual regions for classification and may serve as a general framework to improve the performance of the state-of-the-art methods.

\keywords{Out-of-distribution, Causal Inference, Meta-learning, Invariant feature, Balanced subsets}
\end{abstract}

\section{Introduction}
Deep learning approaches have achieved impressive performance based on the independent and identically distributed (i.i.d.) hypothesis that testing and training data share similar distribution. However, real-world cases may violate the hypothesis due to the complex real data collection or generation mechanism (such as environmental differences \cite{para1-1}, and selection bias \cite{para1-2}). Many studies have revealed the performance of classic machine learning methods has a sharp drop under distributional shifts \cite{OODsurvey,OODsurvey2,para1-3}, which indicates the necessity of learning an excellent model on out-of-distribution (OOD) data. To further promote the development of out-of-distribution generalization research, the NICO Challenge is held.  The NICO Challenge is divided into two tracks: track 1 - public context generalization and track 2 - hybrid context generalization. The context of NICO++ includes two types: common context and unique context. The common context appears in all classes, so it supports the task of track 1, and the unique context only appears in the specific class, which supports the task of track 2.

Commonly used neural networks, such as resnet18 \cite{resnet18} and resnet50 \cite{resnet18}, are difficult to perform well in the OOD data scenarios. To solve this problem, many OOD generalization methods based on causal inference have been proposed \cite{OODsurvey,OODcausal,shen2018,caam}. These algorithms typically aim to capture the invariant causal mechanism and focus on key regions of the image to reduce the impact of contextual diversity factors. CRLR \cite{shen2018} uses feature weighting to make images of different distributions have the same feature distribution. KeepingGood  \cite{keepgood} adopts the idea of staged training to eliminate the adverse effect of data distribution on SGD momentum. CaaM \cite{caam} decouples causal features using CBAM \cite{cbam} attention mechanisms, implements causal interventions based on IRM loss \cite{irm} to obtain cross-domain invariant causal features, and iteratively optimizing data partitioning to prevent excessive intervention. In general, these methods reduce the negative effects of confounding factors and achieve good results. However, we find facts that the subset of training data contains imbalance, which leads to ill-posed learning for causal learners and heterogeneous data leads to weight divergence of multiple classifiers, which may hinder model convergence. These reasons degrade the performance of existing methods.

To address these problems, this paper presents a balanced meta-causal learner (BMCL) that improve the conventional causal learning with balanced task generation and meta learning. The proposed framework follows the CaaM pipelines and it contains two novel modules: the balanced task generation and meta-causal feature learning modules. The balanced task generation module follows CaaM to generate raw data splits and samples meta-tasks therein with balanced three balancing strategies to alleviate the data imbalance problem, including manual balancing, loss balancing, and aggregation balancing with the final partitioning obtained via an aggregation balance algorithm. This balances the class and context of the training images in the tasks to allow the meta-learner to learn the invariant causal features under different visual contexts, and therefore alleviates the ill-posed learning with imbalanced data. The meta-causal feature learning module employs the meta-learning framework to enable the causal learner to retain knowledge learned from all tasks, i.e. data splits, and make it learn quickly to adapt to new tasks and learn unified features. This degrades the model complexity of CaaM by using a single meta-learner, and it fosters the model convengency caused by the weight divergence of its multiple causal learners.

Experiments have been conducted on the subset of the NICO++ dataset and the large datasets of NICO Challenge track1 and track2, including performance comparison, ablation study, and case study. The comparison results verify the generalization ability of the meta causal learner in the OOD case outperforms existing methods. The ablation further illustrates the effectiveness of each component. And case study shows the proposed BMCL can focus on class-invarant visual regions of images in different contexts. To summarize, this paper includes two main contributions: 

\begin{itemize}
    \item A meta causal learner is proposed, which can capture common causal features from different tasks. This can reduce the negative impacts of confounding factors and improve the generalization ability of the model in OOD case.  
    \item It presents a balanced subset partitioning strategy and proves that balanced subsets can enhance the decision-making ability of the model.

\end{itemize}

\section{Related Works}

\subsection{OOD Generalization}

To further promote the research on the problem of agnostic distribution shifts between the training and testing sets, the out-of-distribution(OOD) generalization is proposed for learning a model that performs well under distribution shifts settings \cite{OODsurvey,OODsurvey2,OOD1,OOD2,OOD3}. 

There are OOD generalization problems in many fields, specially in the field of image classification, where the diversity of OOD settings presents challenges: domain adaption \cite{domainsurvey,domain1,domain2},
debiasing \cite{debias1,debias2,debias3},  long-tailed recognition \cite{keepgood,longtail1,longtail2,longtail3,longtail4}. To deal with the OOD problem, \cite{NICO} proposed a real-world OOD dataset NICO, and an extended version of NICO called NICO++ \cite{NICO++} is released for NICO Challenge2022, which has a larger scale with images, contexts and classes. By adjusting the scale of the context, a variety of OOD situations can be simulated, allowing for an in-depth study. 
Domain adaptation tasks \cite{domain1,domain2} find a cross-domain invariant representation by separating task-related and task-independent features. Debiasing tasks \cite{debias2,debias3} improve generalization by training separate models based on data biases to remove biased information. Long-tail classification tasks facilitate classification by building networks to learn feature representations for the majority and minority classes \cite{longtail1}, respectively, or by building a balance of head and tail classes \cite{longtail4}.

\subsection{Causal inference}
Causal inference is an effective method to solve the OOD problem. Pearl \cite{causal} believes that causality is divided into three levels: association, intervention and counterfactuals. Association is the correlation between data obtained by observing data. Intervention is using artificially controlling conditions to reduce the influence of confounding factors, usually through the front door criterion or back door criterion. Counterfactuals speculate on possible outcomes from conditions that did not occur. At present, there are outstanding causal intervention methods, such as re-weighting the samples so that samples in different environments have the same feature space \cite{shen2018,ICP,ICP2,ICP3}. Furthermore, methods using invariant loss can obtain causal features that are invariant cross environments \cite{irm,irm2,irm3,caam}. The counterfactual method \cite{cfc1,cfc2,cfc3} is used to improve the generalization ability and robustness of the model by generating counterfactual samples. In the field of image classification, causal inference aims to make the model focus on the main regions of the image and ignore the impact of the context, that is, to obtain the invariant causal features.

\subsection{Meta Learning}
Meta-learning\cite{metasurvey} is a branch of deep learning that aims to teach models to learn. One of the famous method is MAML \cite{maml}, which is more like a learning strategy than a model and it can be used on other deep network models. Meta-learning builds tasks as the basic unit of training, learns from each task and quickly adapting to the next task. Meta-learning has a wide range of applications: Domain Adaptation and Domain Generalization \cite{meta1,meta2}, in which meta-learning is used to learn regularizer or learn hyperparameters for feature transformation layers, so that the model can adapt to new domains. Another application is Few-Shot Learning \cite{few1,few2}, which uses meta-learning and attention mechanisms to quickly extend to new classes by learning base classes without forgetting knowledge of old classes.

\section{Relation between BMCL, CBAM and CaaM}

\subsection{From Attentive Feature Learning to Causal Learning}
OOD-distributed images amplify the impact of context on classification by creating pseudo-associations between context and categories, which requires the model to better focus on the subjects in images. To this goal, the work of CBAM \cite{cbam} designed a lightweight attention module to focus on the class-predictive features $\textbf{F}_{att}$, which is learned on the basis of raw features extracted by visual modal $V(.)$, i.e., $\textbf{F}_{att}=CBAM(V(x))$, where $x$ is the images. 

The effectiveness of the CBAM is limited because of the attentive bias caused by wrong associations between contextual information and classes. To addressed this, CaaM \cite{caam} improves the networks and the training pipeline to learn the context-irrelevant visual representation, as shown in Fig. \ref{fig1} (a). CaaM first learns to decouple causal features $\textbf{F}_{c}$, confounding features $\textbf{F}_{s}$, and mix features $\textbf{F}_{x}$ from images, that is, $\textbf{F}_{c}, \textbf{F}_{s}, \textbf{F}_{x}=CaaM(x)$. Then CaaM generates training partitions $\mathcal{T}=\{t_1,t_2,\cdots,t_m\}$ based on contextual information, and applies multiple causal learner for aggregating the knowledge of models.

\subsection{The Enhanced Causal Learning by BMCL}
\begin{figure}[t]
\includegraphics[width=1\linewidth]{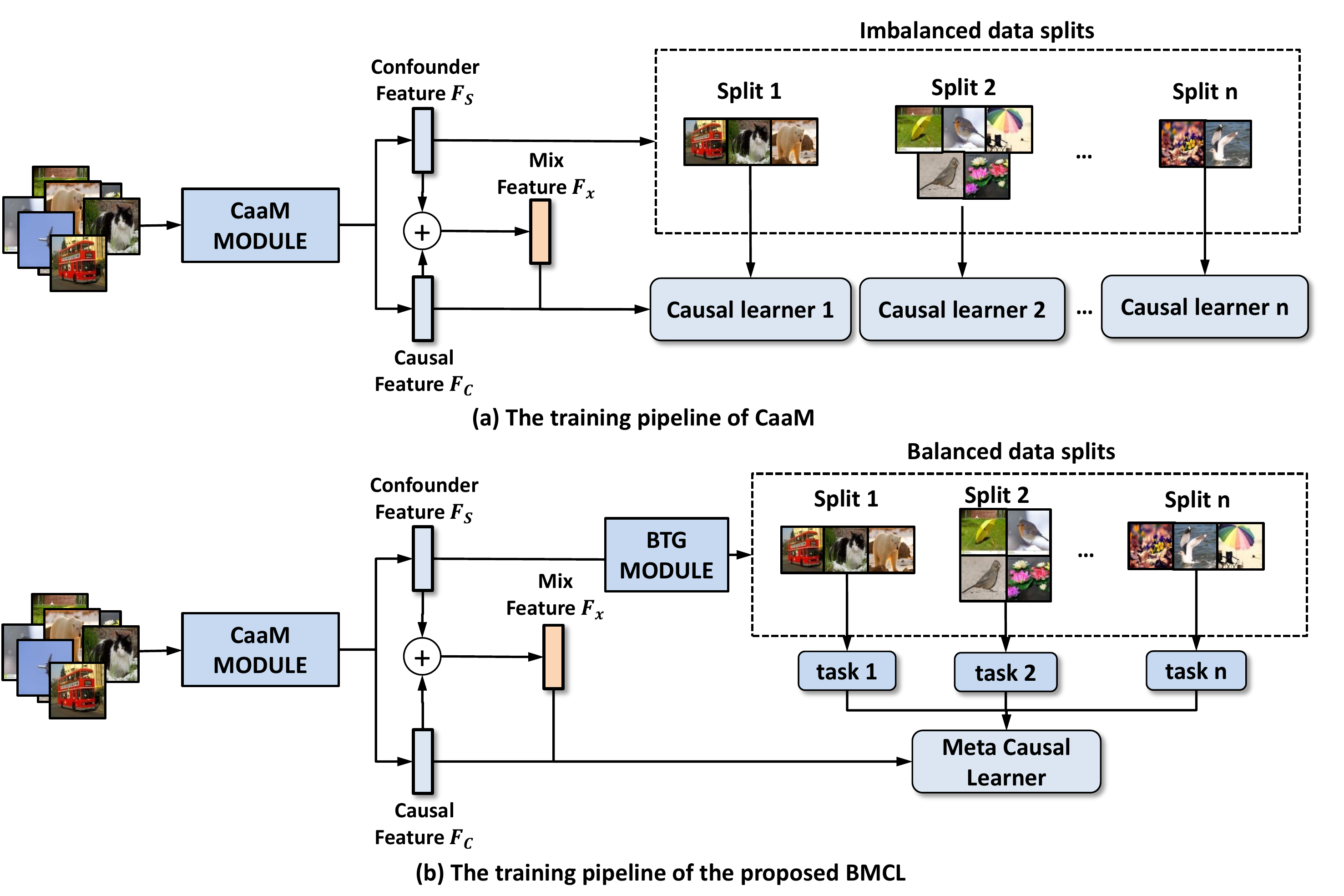}
\caption{Illustration of the training pipeline of CaaM and the proposed BMCL. BMCL makes the data partitioning more balanced, and adapts meta learner to enhance the effects of learning invariant features from different source data.} \label{fig1}
\end{figure}

As mentioned above, CaaM proposes to aggregate the knowledge learned in different partitions to obtain a better predictor, but the imbalance of partitions brings about mutual interference during aggregation, and the learning in different partitions needs to be more autonomous. The proposed BMCL addresses these problems and the training pipeline is shown in Figure \ref{fig1} (b).

BMCL follows the feature decoupling in CaaM, obtains $\textbf{F}_{c}, \textbf{F}_{s}, \textbf{F}_{x}$, and designs the BTG module to get the balanced partition $\mathcal{T}'=\{t'_1,t'_2,\cdots,t'_m\}$. In each partition, the learning is improved by means of setting corresponding task, and a meta-learner is used to learn unified class-level features, thereby enhancing the effects of causal learning.

\begin{figure}[t]
\includegraphics[width=1\linewidth]{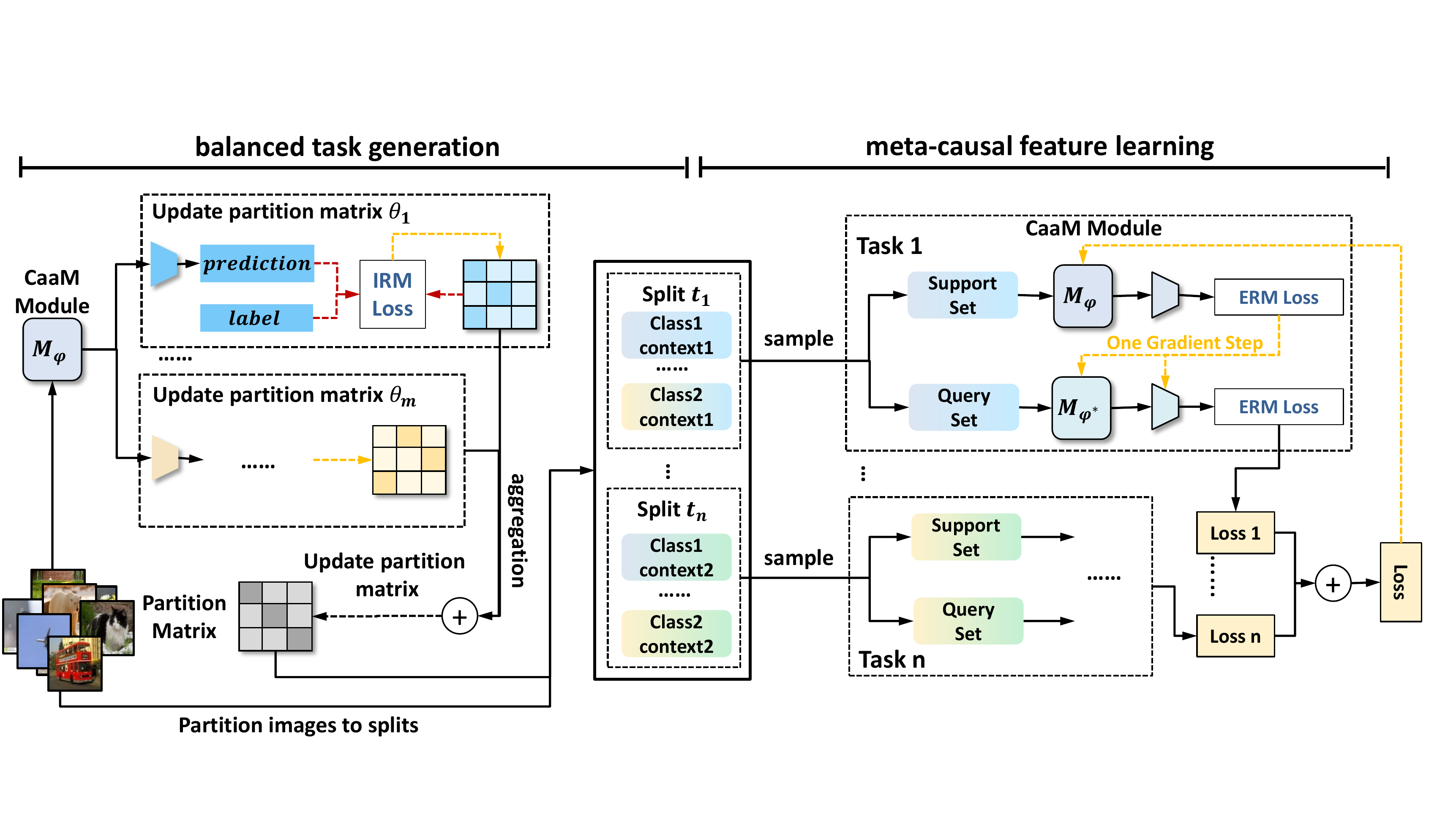}
\caption{Schematic diagram of the BMCL algorithm: the balanced task generation module divide the images into balanced splits with different contexts. The meta-causal feature learning module samples images from different partition to generate tasks, and learn invariant causal features from different tasks.} \label{fig2}
\end{figure}

\section{Our approach}
\subsection{Overview}
As shown in Figure \ref{fig2}, the proposed BMCL has two main modules: balanced task generation (BTG) module and meta-causal feature learning (MCFL) module. The BTG module uses the aggregation balance method to train multiple partition matrices. This can divide the dataset into multiple subsets with different contexts in a balanced manner. The MCFL module aims to learn invariant causal features to reduce the negative effects of confounding features. This enables the model to focus on the main regions of the image with different contexts.

\subsection{Balanced task generation (BTG) module}
The BTG module uses the confounder features $\textbf{F}_s$ extracted by the CaaM to train the partition matrix $\theta$ and update the data partition $\mathcal{T}=\{t_1,t_2,\cdots,t_m\}$. BTG first uses a random partition matrix. To get the updated partition matrix, a bias classifier $h(\cdot)$ is trained for each matrix and the prediction $c = h(\textbf{F}_s)$ can be obtained. For this, the BTG module can distinguish the contexts and it is optimized by miniming the ERM Loss, i.e.
\begin{equation}
\mathcal{L}_{bias}^{erm} = E_{(x,y)\in \mathcal{D}}\ell(h(F_s),y)
\end{equation}
where $\mathcal{D}$ is all training data, $y$ is the label. Then using this classifier, under the constraint of an IRMloss \cite{irm}, a partition matrix $\theta$ is trained gradually. We maximize the loss to make sure the difference among all splits. After training we update the partition of the dataset in a fine-grained way:
\begin{equation}
\mathcal{L}_{split}^{irm} = \sum_{t\in\mathcal{T}_i(\theta)} R^t(h) + \lambda \cdot {\Vert \nabla_{w\mid w=1.0} R^t(w\cdot h)\Vert}^2
\end{equation}
where $\mathcal{T}=\{t_1,t_2,\cdots,t_m\}$ is current data partition, $\mathcal{T}_i(\theta)$ denotes partition $\mathcal{T}_i$ is decided by $\theta \in \mathcal{R}^{K\times m}$, $K$ is the total number of training samples and $m$ is the number of splits in a partition, $R^t(h):=E_{(x,y)\in t_i}\ell(h(F_s),y)$ is the loss under subset $t_i$, $h(\cdot)$ is the bias classifier trained in the previous step, $w = 1.0$ is a scalar and fixed “dummy” classifier, the gradient norm penalty is used to measure the optimality of the dummy classifier at each subset $t$, and $\lambda \in [0,\infty]$ is a regularizer balancing weight between the ERM term and the invariance of the predictor $1 \cdot h$.

During this process, we found training only one partition matrix $\theta$ may lead to a large imbalance in subsets, which may lead to a poor performance of the model. In order to alleviate the imbalance, BMCL employs three balancing strategies and discusses this in the ablation study. Eventually, BMCL uses multiple matrices to decide the final partition in BTG module. It combines the probability distributions of multiple training to alleviate the problem of imbalance. And then the aggregation method is used to obtain the final partition, i.e.
\begin{equation}
\theta_{final} = \sum_{i=0}^n \begin{pmatrix}
        p(1,1) & \cdots & p(1,m) \\
        \vdots & \ddots & \vdots \\
        p(k,1) & \cdots & p(k,m)
    \end{pmatrix}_i
\end{equation}
where $\theta_{final} \in R^{K\times m}$, $p(k,m)$ denotes the probability that the $k^{th}$ image is divided into the $m^{th}$ partition.
For $\theta_{final}$, the index of the split to be divided into is:
\begin{equation}
Idx = \mathop{argmax}\limits_{\theta}(Softmax(\theta_{final}))\label{div}
\end{equation}
Then, dividing the $K$ images into corresponding data subsets based on Equation \ref{div}.  The BTG module divides the dataset into fine-grained subsets with different contexts, which helps the model adapt to other tasks and extract cross-domain invariant causal features.

\subsection{Meta-causal feature learning (MCFL) module}
Commonly used a way to solve the OOD problem of image classification is to extract causal features of image subjects from different contexts. There is a feasible solution that is to use the properties of meta-learning cross-task learning to extract invariant features from different tasks.
Based on the idea of meta-learning, the MCFL module first organizes the training set into the form of a task, which consists of a support set and a query set for the training process. For $m$ splits, we organize $m$ tasks as an update. Each task is from one split. The total number of meta-tasks is $s$, which can be divided by $m$. The composition of the task is to randomly sample $w$ classes, and then randomly sample $i+j$ images from each class, of which $i$ images for the support set and $j$ images for the query set.

During the training phase, each meta-task has its own independent training and testing sets, called support and query sets. The MCFL first promotes fast convergence of the model by simply updating the support set training in the meta-training process.
The model first learns a parameter from the support set and then updates the model. Note, we define the feature extractor as $M_\varphi$ with parameters $\varphi$ and the classifier as $f_\mu$ with parameters $\mu$. We use the feature extractor $M_\varphi$ and classifier $f_\mu$ to get the causal feature $F_c$ and then cauculate an ERM Loss for the first model updating:
\begin{equation}
F_c = M_\varphi(x_{spt})
\end{equation}
\begin{equation}
\mathcal{L}_{\mathcal{K}_s} = \ell(f_\mu(F_c),y_{spt})
\end{equation}
where the $x_{spt}$ and $y_{spt}$ is the images and labels in support set of task $\mathcal{K}_s$, and the $\ell$ is the cross-encropy loss.
The updated parameter vector $\varphi^*$ and $\mu^*$ are updated using one or multiple gradient descent on the current support set calculated by $\mathcal{L}_{\mathcal{K}_s}$.  For example, when using one gradient update:
\begin{equation}
\varphi^* = \varphi - \alpha\nabla_{\varphi}\mathcal{L}_{\mathcal{K}_s}
\end{equation}
\begin{equation}
\mu^* = \mu - \alpha\nabla_{\mu}\mathcal{L}_{\mathcal{K}_s}
\end{equation}
where the step size $\alpha$ may be fixed as a hyperparameter or meta-learned.

For meta-learning, a good meta learner should perform well on all meta tasks. Therefore, BMCL has only one meta learner and uses ERM loss to update this single learner.
We use the updated model to train on the query set and get the training loss. Since the support set and the query set have the same classes, after learning on the support set to get $\varphi^*$, the learning of causal features can be promoted on the query set, and show a good performance. For building meta-task from splits, the updated model can be obtained after $m$ tasks:
\begin{equation}
F_c = M_{\varphi^*}(x_{qry})
\end{equation}
\begin{equation}
\mathcal{L}_{train}^{meta} = \frac{1}{m}\sum_{i=0}^m \mathcal{L}_{\mathcal{K}_i} = \frac{1}{m}\sum_{i=0}^m \ell(f_{\mu^*}(F_c),y_{qry})
\end{equation}

It is worth noting that the meta-optimization is performed over the model parameters $F_\varphi$, whereas the objective is computed using the updated model parameters $\varphi^*$.
\begin{equation}
\varphi \xleftarrow[]{} \varphi - \beta\nabla_{\varphi} \sum_{i=0}^m \mathcal{L}_{\mathcal{K}_i}
\end{equation}
\begin{equation}
\mu \xleftarrow[]{} \mu - \beta\nabla_{\mu} \sum_{i=0}^m \mathcal{L}_{\mathcal{K}_i}
\end{equation}
where $\beta$ is the meta step size. By training on multiple tasks and continuously minimizing the sum of losses on all tasks, the model can accurately extract causal features. This provides a guarantee for the model to achieve satisfactory performance on new tasks. In addition, a ERM loss is used to enhance the invariance of features, in this stage, we use mixup \cite{mixup} to strengthen the images:
\begin{equation}
\begin{gathered}
\Tilde{x} = \lambda x_i + (1-\lambda)x_j\\
\Tilde{y} = \lambda y_i + (1-\lambda)y_j
\end{gathered}
\end{equation}
And the loss was calculated as:
\begin{equation}
F_c = M_\varphi(\Tilde{x})
\end{equation}
\begin{equation}
\mathcal{L}_{train}^{erm} = \ell(g(F_c),\Tilde{y})
\end{equation}
where the $g(\cdot)$ is a classifier for this process alone.

\subsection{Training Strategy}

BMCL incorporates two stages model training process:
\begin{itemize}
    
    \item  \textbf{Stage 1:} Training balanced subsets partition. The initial partition matrix is randomly generated. And then a balanced constrain is used to update the partition matrices. First, we train a biased classifier using the empirical risk loss $\mathcal{L}_{bias}^{erm}$ and use the classifier to constrain the data partition by an invariant risk loss $\mathcal{L}_{split}^{irm}$. Significantly, we minimize the empirical risk loss $\mathcal{L}_{bias}^{erm}$, but maximize the invariant risk loss $\mathcal{L}_{split}^{irm}$, i.e.

    \begin{equation}
    \mathop{min} \mathcal{L}_{bias}^{erm} - \lambda \mathcal{L}_{split}^{irm}
    \end{equation}

    \item  \textbf{Stage 2:} The extraction of training features is jointly constrained by the empirical risk loss from different tasks and from the whole dataset. In order to obtain robust features, we minimize the loss:
    \begin{equation}
    \mathop{min} \mathcal{L}_{train}^{meta}+\mathcal{L}_{train}^{erm}
    \end{equation}
    
\end{itemize}

\section{Experiments}
\subsection{Datasets}
\subsubsection{NICO++} \cite{NICO++}. A real-world image classification dataset under OOD settings, that is, the contexts of images in the testing may be unseen during training. To achieve this, NICO++ decomposes images into subject concepts and visual contexts, so that the contextual distribution in training and testing can be easily adjusted. There are typically two types of contextual setting in NICO++, namely, the common context setting (as in track 1 of NICO++ Challenge) and the hybrid context setting (as in track 2 of NICO++ Challenge). The common contexts setting means that all classes share identical contexts both in training and testing; and the unique contexts setting means each class has unique contexts. In details, NICO++ currently includes 200,000 images in 80 categories, ranging from animals, plants, traffic to objects; images in each category are organized into 10 public contexts and 10 unique contexts.
\subsubsection{Track1}. A subset sampled from NICO++ under common context setting. The track1 dataset has 88,866 images for training, 13,907 for public testing, and 35,920 for private testing. All contexts in track1 is existing in all categories.
\subsubsection{Track2}. A subset sampled from NICO++ under hybrid context setting. The dataset for track2 has 57,425 images for training, 8715 for public testing, and 20,079 for private testing. in track2, each category includes some unique contexts, which makes the situation more complected.
\subsubsection{NICO++(subset)}. To further investigate the ability of OOD generation, we used a self-made datset termed as NICO++(subset) in experiments. In details, sampled images contain 6 contexts in 10 classes including animals, vehicles, and others from the NICO++. Furthermore, we follow the long-tailed and zero-shot settings in previous work \cite{caam}, and additionally set 4 contexts for training and 2 contexts unseen during training for testing, the size of training, validating, and testing are 2,870, 1,754, and 1755  respectively.

\subsection{Evaluation Protocol}
We use the Top-1 Accuracy on the validation set and test set for evaluation as the previous work \cite{caam} of causal learning did. The formula is as follows:
\begin{equation}
    Accuracy=\frac{TP+TN}{TP+TN+FP+FN}
\end{equation}
where TP, TN, FP, FN are the number of true positive, false positive samples, true negative samples, and false negative samples.

\subsection{Implementation Details}
For experiments in all the datasets, we used SGD as the optimizer, and setting learning rate from 0.01 to 0.03. Models were trained for 220 epochs and the learning rate was decayed by 0.1 at 200 epoch. In settings of partition training in CaaM and BMCL, the number of partition is 4 and from the 40-th epoch, the data partitions were updated every 20 epochs. For the process of updating splits, we followed the previous paper and trained each for 100 epochs with early stopping, when epoch is over 40 and accuracy no longer increases more than 5 epoch. The optimizer for this process was set to SGD with learning rate as 0.1. And the $\lambda$ for IRM Loss was set to 1e6. For the meta-task, we choose 3 classes for one task, and 1 or 2 image(s) for the support sets, 11~15 images for the query set. The updated learning rate of the one step gradient of meta was set to 0.005~0.01. For the erm training process, we set the batch size to 32*4.

\subsection{Performance Comparison}

This section shows the performance comparison of BMCL with other image classification methods. We typically divided them into visual learning based on convolutional networks (Conv. Method) including the traditional ResNet-18 \cite{resnet18}, the widely-used data augmentation methods Mixup \cite{mixup}; and methods using causal inference, including the CBAM \cite{cbam} and the CaaM \cite{caam}. The drawn following observations from Table \ref{tab:comparison}:
\begin{itemize}
    \item Generally speaking, BMCL achieved better performance than other algorithms in all cases. It is reasonable since BMCL is able to generate balanced subsets to away from ill-posed learning.
    Benefiting from this property, BMCL typically outperformed the other methods on both datasets with a large margin up to 12\%.  

    \item Basic visual networks such as ResNet-18 are prone to bias the context and lead to poor predictions when faced with OOD images. The Mixup enhancement enriches the context type corresponding to the category subject to a certain extent through image interpolation during training, thereby implicitly enhancing the model's attention to the subject part, and achieving an improvement of about 8\% on the small data set NICO++ (subset). On the larger datasets Track 1 and Track 2, due to the richer contextual information, the improvement is also more obvious by 13\% and 19\%.

    \item Methods based on causal learning generally perform better than traditional vision networks on large datasets, mainly because they design explicit methods to pay attention to the predictive features of images, and the improvement is more stable in the presence of sufficient data.

    \item However, on the small data set NICO++ (subset), the method based on causal learning did not bring significant improvement to the results. For CBAM, the attention mechanism is easy to focus on the background on a small amount of data, which aggravates the impact of OOD on model prediction. However, the update of CaaM is prone to imbalance. In a small data set, some predictors only have few samples, which is likely to have a bad impact on the overall prediction.

\end{itemize}

\begin{table}[t]
\centering \caption{Classification accuracies (\%) of algorithms on track1, track2, and NICO++(subset) datasets using ResNet18 as backbone. "val" and "test" denote the accuracies on validating set and testing set. "pub test" denote the accuracies on the publish testing set of Nico++ Challenge.}
\label{tab:comparison}
\centering
\setlength{\tabcolsep}{2mm}{
\begin{tabular}{|c|c|cc|c|c|}
\hline
\multirow{2}{*}{Method} & \multirow{2}{*}{Model} & \multicolumn{2}{c|}{NICO++(subset)} & NICO++-track1 & NICO++-track2 \\ \cline{3-6} 
 &  & \multicolumn{1}{c|}{val   acc} & test   acc & pub   test & pub   test \\ \hline
\multirow{2}{*}{\begin{tabular}[c]{@{}c@{}}Conv.\\ Method\end{tabular}} & ResNet-18 & \multicolumn{1}{c|}{44.73} & 45.93 & 58.07 & 47.11 \\ \cline{2-6} 
 & Mixup & \multicolumn{1}{c|}{49.00} & 49.06 & 66.47 & 56.25 \\ \hline
\multirow{3}{*}{\begin{tabular}[c]{@{}c@{}}Causal\\  Method\end{tabular}} & CBAM & \multicolumn{1}{c|}{44.27} & 45.47 & 65.35 & 58.71 \\ \cline{2-6} 
 & CaaM & \multicolumn{1}{c|}{43.93} & 46.44 & 72.93 & 66.44 \\ \cline{2-6} 
 & BMCL & \multicolumn{1}{c|}{\textbf{51.45}} & \textbf{52.02} & \textbf{84.66} & \textbf{78.81} \\ \hline
\end{tabular}}
\end{table}

\subsection{Ablation Study}
\begin{table}[t]
\centering \caption{Results of the ablation study, showing the effectiveness of each module in BMCL on classification performance.}
\label{tab:ablation}
\centering
\setlength{\tabcolsep}{8mm}{
\begin{tabular}{|c|cc|}
\hline
\multirow{2}{*}{Model} & \multicolumn{2}{c|}{NICO++(subset)} \\ \cline{2-3} 
 & \multicolumn{1}{c|}{val   acc} & test   acc \\ \hline
Baseline & \multicolumn{1}{c|}{44.73} & 45.93 \\ \hline
+CBAM & \multicolumn{1}{c|}{44.27} & 45.47 \\ \hline 
+CBAM+Causal & \multicolumn{1}{c|}{43.93} & 46.44 \\ \hline
+CBAM+Causal+BTG & \multicolumn{1}{c|}{46.38} & 47.46 \\ \hline
+CBAM+Causal+meta & \multicolumn{1}{c|}{49.74} & 49.52 \\ \hline
+CBAM+Causal+BTG+meta & \multicolumn{1}{c|}{\textbf{51.45}} & \textbf{52.02} \\ \hline
\end{tabular}}
\end{table}

In this section, we investigate the effectiveness of the proposed algorithm. The experiment selected resnet18 \cite{resnet18} as the baseline.
Our BMCL method is divided into two modules, a balancing module and a meta-learning module.  As can be seen from Table \ref{tab:ablation}, we tested two modules based on CaaM respectively, and both achieved good results.  Then we use the two modules at the same time, and use the meta-model to learn causal features while ensuring the balance of the dataset partition, and achieve the current best performance.

\subsection{In Depth Analysis of Balancing Strategy and Meta Learner}
\begin{table}[t]
\centering \caption{The results of different combination of feature learning and balancing strategy.}
\label{tab:balance}
\centering
\setlength{\tabcolsep}{5.8mm}{
\begin{tabular}{|c|c|cc|}
\hline
\multirow{2}{*}{Feature   learning} & \multirow{2}{*}{Balancing strategy} & \multicolumn{2}{c|}{NICO++(subset)} \\ \cline{3-4} 
 &  & \multicolumn{1}{c|}{val   acc} & test   acc \\ \hline
\multirow{4}{*}{Backbone} & - & \multicolumn{1}{c|}{43.93} & 46.44 \\ \cline{2-4} 
 & LB & \multicolumn{1}{c|}{46.04} & 47.24 \\ \cline{2-4} 
 & MB & \multicolumn{1}{c|}{44.05} & 47.41 \\ \cline{2-4} 
 & GB & \multicolumn{1}{c|}{\textbf{46.38}} & \textbf{47.46} \\ \hline
\multirow{4}{*}{Meta   learner} & - & \multicolumn{1}{c|}{49.74} & 49.52 \\ \cline{2-4} 
 & LB & \multicolumn{1}{c|}{49.86} & 50.26 \\ \cline{2-4} 
 & MB & \multicolumn{1}{c|}{\textbf{51.62}} & 49.57 \\ \cline{2-4} 
 & GB & \multicolumn{1}{c|}{51.45} & \textbf{52.02} \\ \hline
\end{tabular}}
\end{table}

Table \ref{tab:balance} shows the performance of using different balancing methods, including Loss Balance(LB), Manual Balance(MB), Aggregation Balance(GB). Among them, LB is to add loss in the process of training the partition matrix, it can automatically balance the number of images in different subsets during the training process.  MB is to manually select the images of each class when dividing the splits, and follow the principle of more deletion, less complement, and move the extra images from a split to other splits. GB is a smooth operation that reduces the chance and extremeness of the partition by training multiple partition matrices, so as to alleviate the imbalance.

By testing different balancing methods, we found that GB is the best, and the reason why LB and MB are not good may be because the forced full balancing leads to destroying the structure of data partitioning, which is disadvantageous to improve performance. While GB can alleviate the imbalance, it can retain the structure of data partition. As observed, GB balance method achieved the best results both under CaaM and under the meta learner and outperformed the other on NICO++(subset) with 2\%.

\subsection{Case Study}
\begin{figure}[t]
\includegraphics[width=1\linewidth]{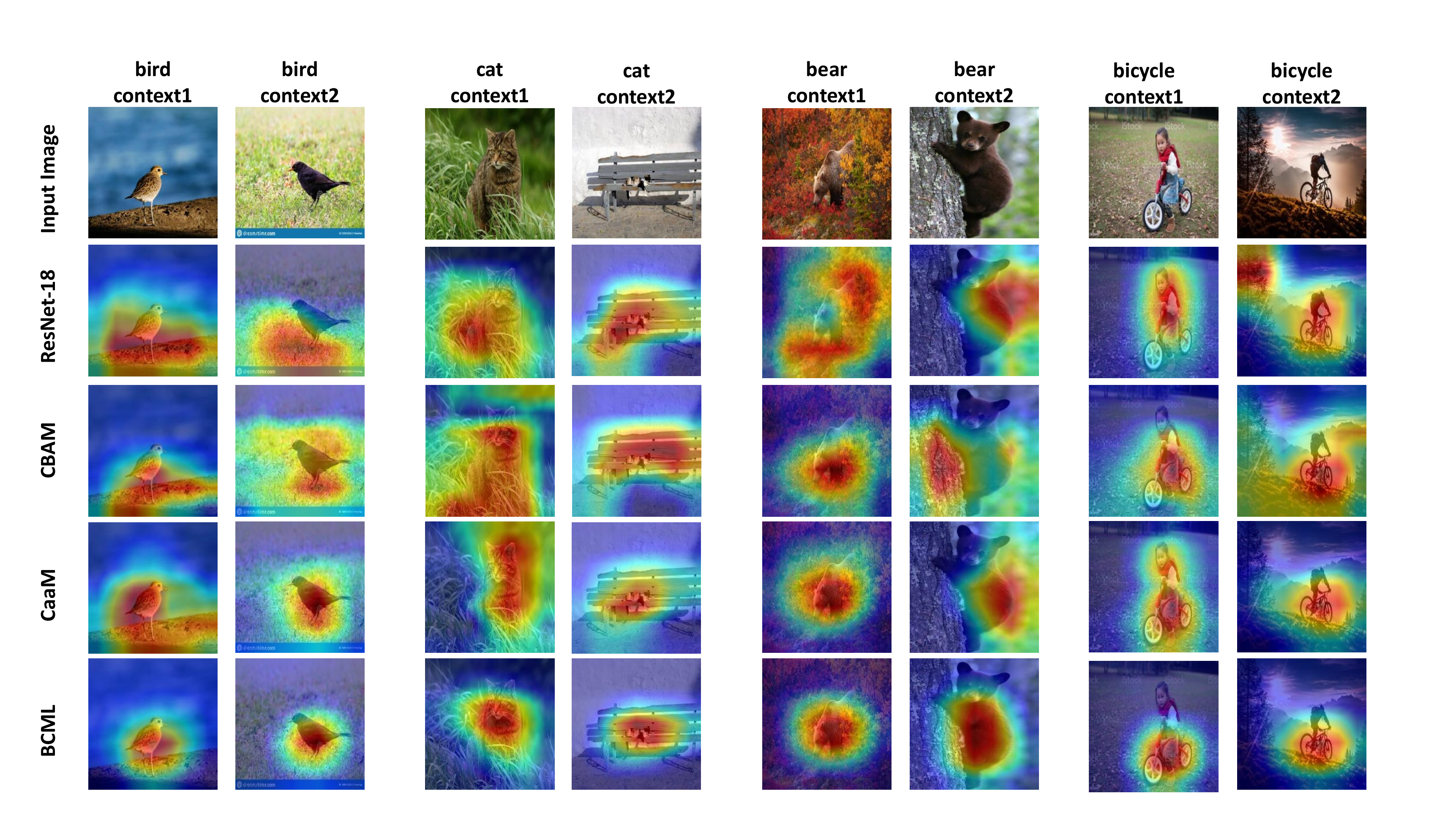}
\caption{Visualization of attention maps with base model, CBAM, CaaM, and BMCL} \label{fig3}
\vspace{-0.3cm}
\end{figure}

Figure \ref{fig3} shows the attention maps generated by the proposed BMCL, the commonly-used backbone ResNet18, the attention mechanism CBAM, and the causal inference method CaaM. This is achieved by using the Grad-CAM \cite{gradcam}. In this experiment, four classes images from the NICO++ test dataset are selected and each class of images contains two contexts. As observed, BMCL can focus on the class meaningful regions of the image rather than the contexts. For the ResNet18, it can easily be influenced by the surrounding environment and focus on meaningless regions, such as a cyclist and a light in the background. For the CBAM, it always pays attention to a wide range, this contains the important regions and a large of noisy regions besides.   
The CaaM achieves the good performance compare with ResNet18 and CBAM, which indicates the reliability of causal inference. But CaaM may learn biased feature attention maps, such as a cyclist in bicycle image and a bench in cat image. This could be caused by unbalanced data learning. Significantly, BMCL achieves the best performance. It can focus precisely on the class-invarant visual regions, and exclude the interference of complex background. These observations verify the effectiveness of BMCL for invariant causal feature learning.

\section{Conclusions}
This paper presents a novel approach, termed BMCL, to cope with the challenge of agnostic distribution shifts in out-of-distribution (OOD) settings, which can perform a self-learning balanced subset partition method to generate balanced subsets and learn the invariant causal features based on meta-learning. Notably, BMCL can keep the causal learner away from ill-posed learning and reduce the model complexity. Experimental results show that BMCL can alleviate the interference of confounder factors and enhance the learning and generalization ability of the model in the OOD case.

This study can be further explored in two directions. First, a cost-effective subsets partition method can be explored to reduce the time cost, such as combining curriculum learning to partially sample the data. Second, BMCL can be extended to more challenging settings, such as Federated Learning. \\
\textbf{Acknowledgments} This work is supported in part by the Excellent Youth Scholars Program of Shandong Province (Grant no. 2022HWYQ-048) and the Oversea Innovation Team Project of the  "20 Regulations for New Universities" funding program of Jinan (Grant no. 2021GXRC073)



%
%
\bibliographystyle{splncs04}
\bibliography{egbib}
\end{document}